\documentclass[letterpaper, 10 pt, conference]{ieeeconf}  

\IEEEoverridecommandlockouts                              

\usepackage{graphics} 
\usepackage{amsmath} 
\usepackage{amssymb}  
\usepackage{cite}
\usepackage[ruled,vlined]{algorithm2e}
\usepackage{tikz}
\usetikzlibrary{arrows,shapes,snakes,automata,backgrounds,petri}

\newcommand{\rd}{{\mathrm d}}

\newcommand{\rT}{{\mathrm{T}}}

\newcommand{\vx}{{\bf x}}

\newcommand{\vz}{{\bf z}}

\newcommand{\vq}{{\bf q}}

\newcommand{\vf}{{\bf f}}

\newcommand{\vu}{{\bf u}}
\newcommand{\vv}{{\bf  v}}

\newcommand{\vF}{{\bf F}}

\newcommand{\argmin}{\operatornamewithlimits{argmin}}

\newcommand{\calJ}{\mathcal{J}}

\newcommand{\calA}{\mathcal{A}}
\newcommand{\calU}{\mathcal{U}}

\newcommand{\Rb}{\mathbb{R}}

\newcommand{\pluseq}{\mathrel{+}=}

\newtheorem{definition}{Definition}

\title{CS-8751}

\title{\LARGE \bf
Autonomous Racing with AutoRally Vehicles and Differential Games}

\author{Grady Williams, Brian Goldfain, Paul Drews, James M. Rehg, and Evangelos A. Theodorou 
\thanks{The authors are with Institute for Robotics and Intelligent Machines at the Georgia Institute of Technology, 
Atlanta, 
GA, USA. 
Email: gradyrw@gatech.edu}
}

\begin{document}

\maketitle
\thispagestyle{empty}
\pagestyle{empty}

\begin{abstract}
Safe autonomous vehicles must be able to predict and react to the drivers around them. Previous control methods rely heavily on pre-computation and are unable to react to dynamic events as they unfold in real-time. In this paper, we extend Model Predictive Path Integral Control (MPPI) using differential game theory and introduce Best-Response MPPI (BR-MPPI) for real-time multi-vehicle interactions. Experimental results are presented using two AutoRally platforms in a racing format with BR-MPPI competing against a skilled human driver at the Georgia Tech Autonomous Racing Facility.
\end{abstract}
\section{INTRODUCTION}\label{intro}

Autonomous vehicles operating in the real world will have to interact with human controlled vehicles. To become competent and safe drivers this means that autonomous vehicles must predict the actions of human drivers and reason about how their own actions might effect the actions of other drivers. Particularly interesting is the case of aggressive driving, which is when the necessary interactions are highly dynamics (e.g. weaving traffic on a freeway, collision avoidance), since the solution must computed quickly while incorporating the low-level dynamic constraints of both vehicles.

\begin{figure}[t]
\includegraphics[width = \columnwidth]{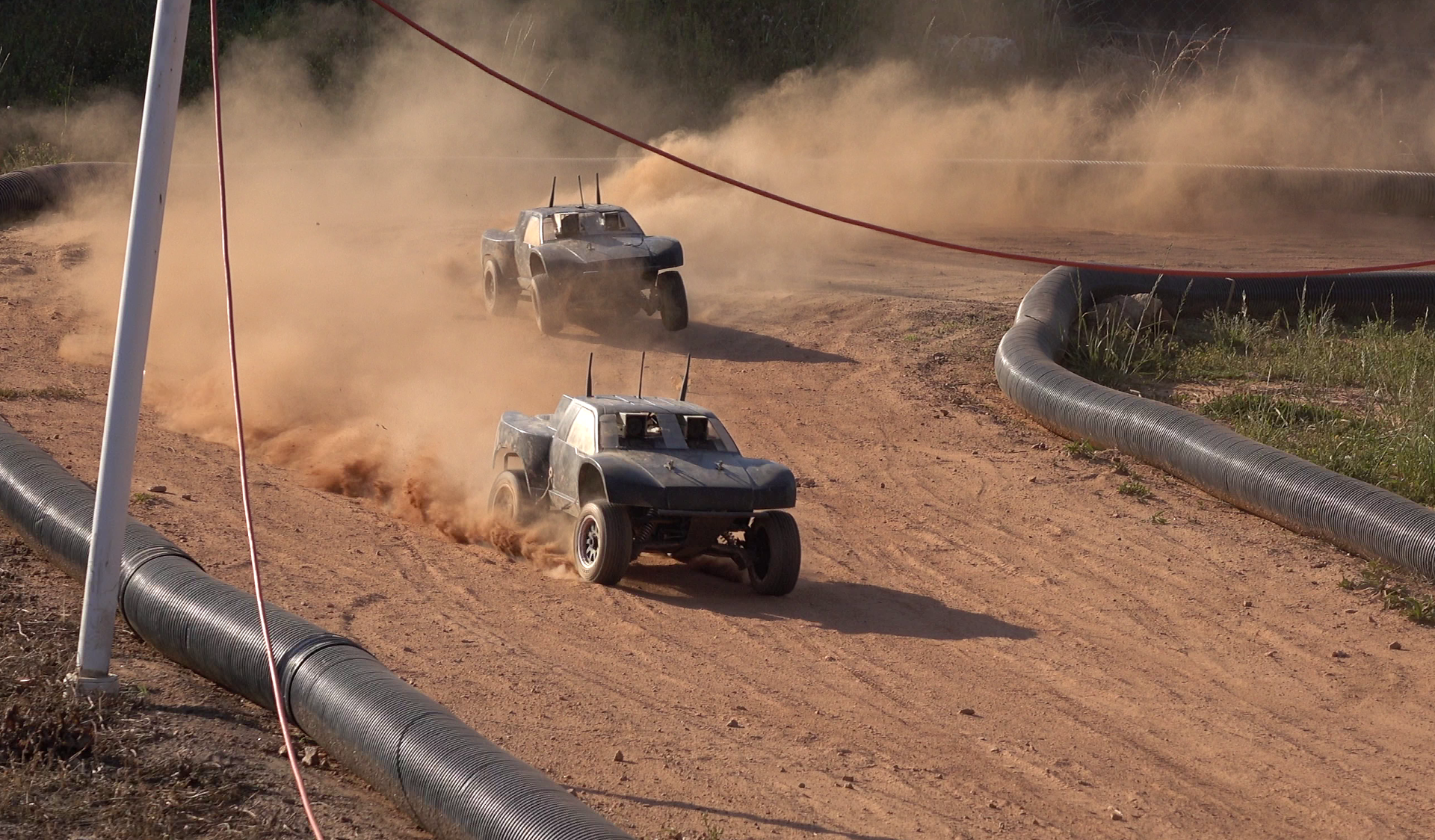}
\caption{Two one-fifth scale AutoRally vehicles racing on the test track at the Georgia Tech Autonomous Racing Facility.}
\label{Fig:TwoPlatforms}
\end{figure}

As a model system for aggressive driving, we are interested in the problem of autonomous racing between two or more vehicles. Autonomous racing necessarily involves pushing the vehicle to its handling/acceleration limits, and there is only a small margin of error when racing against a capable adversary. These factors make autonomous racing a good surrogate problem for studying aggressive driving. As racing platforms we use the Georgia Tech AutoRally platform (Fig. \ref{Fig:TwoPlatforms}) . These are 1/5 scale AutoRally trucks that carry a full desktop computer on-board an are capable of speeds exceeding 50 mph, they are useful stand-ins for full-sized vehicles since they are large enough to exhibit complex non-linear dynamics similar to full-sized vehicles, yet small enough to be robust to repeated crashing.

Autonomously racing against a human pilot is difficult from both an algorithmic and experimental point of view. Algorithmically, the problem of autonomous driving in the presence of other intelligent agents falls into a class of problems known as differential games \cite{isaacs1999differential}. Differential games are notoriously difficult to solve, especially for non-linear systems subject to complex costs and constraints. From an experimental point of view testing aggressive interactions between autonomous vehicles and human controlled ones is problematic with full sized vehicles, since safety becomes a primary concern. This means that these types of interactions have previously only been studied in simulation. In this paper, we overcome these challenges and present an algorithmic and experimental framework for autonomous racing on real robotic platforms. Our mathematical framework for approximately solving differential games is based on combining a sampling based model predictive control algorithm with the algorithmic concept of best-response dynamics from game theory, and we test this algorithmic approach in the real world against a human pilot using two one-fifth scaled AutoRally vehicles modified for vehicle to vehicle communication in order to share relative poses between the two vehicles.

\section{RELATED WORK}

This is the first work which considers autonomous racing against a human adversary in a real world setting. However, there are a number of related works in robotics dealing with both differential game theory, and autonomous driving testbeds.

\subsection{Differential Games}

The theory of differential games was originally studied by Isaacson in \cite{isaacs1999differential} where he laid out the mathematical connections between control theory, optimization, and game theory. In particular, the Hamilton-Jacobi-Isaacs partial differential equation provides a way to solve for the value function of certain types of differential games. However, this is not, and was not intended to be, a practical way to solve general differential games due to it suffering from the curse of dimensionality. Although this seminal work is still relevant because it defined the mathematical problem precisely, it does not contain a practical solution method to the problem of autonomous racing.

The application of the theory of differential games for min-max problems is popular in robotics. Application areas fit into two areas: single player games where system noise is considered adversarial, and pursuit evasion games. An example of the former is \cite{morimoto2003minimax} where the authors use a min-max formulation of a model-based reinforcement learning approach, based on differential dynamic programming, to control a bipedal walker. The idea is that the modeling error is treated as an adversary (a mathematical formulation of Murphy's law) which forces the controller to plan for the worst case scenario. Although this work shares similarities to our racing problem, it does control a complex dynamical system with a learned model, its formulation as a differential game is fundamentally different from our racing problem. The key difference is that the control and adversary signals enter the system through the same input channels, whereas in our racing case the controller and the adversary control entirely different systems.

The latter approach, using min-max for pursuit evasion games, is demonstrated by \cite{walrand2011harbor}. In this work, the authors formulate harbor defense as a pursuit evasion game where the ``defender'' tries to deny harbor access to an attacking agent. The goal of the defender is to close to within distance $\delta$ of the attacker (in order to destroy it) and the goal of the attacker is to get into the harbor while staying distance $\delta$ away from the attacker. These agents are modeled using simple tricycle dynamics with a constant speed, and optimization is performed using the SNOPT solver \cite{gill2005snopt} to perform model predictive control. The difficulty with this approach is getting the solver to return a feasible solution, and even with simple dynamics the authors have to perform a transformation on the dynamics to consistently get a feasible solution. Although this approach could theoretically be applied to our racing problem, the dynamics of our vehicles are much more complicated than what is considered in \cite{walrand2011harbor}, so we should not expect it to be successful. Another similar example is \cite{pan2012pursuit}, where pursuit evasion problems in a 2-D plane are considered. However, the dynamics considered in \cite{pan2012pursuit} are simple single integrator dynamics and therefore not suitable to our problem.

The most similar framework to the one we present here is in \cite{sadigh2016planning}, where a best-response type iteration is used in order to study the effect of robot actions on human behavior in a variety of normal driving scenario. In this work we focus on applying best-response iteration for two vehicles pushed to their performance limits.

\subsection{Autonomous Driving Platforms}

Experimental results for autonomous vehicle research normally rely on custom built platforms. The current round of autonomous vehicle interest and development was spurred largely by the DARPA Urban Challenge \cite{buehler2009urban} where university and industry teams brought autonomous vehicles to compete on a closed course in realistic driving scenarios. Most robots that competed were modified commercial vehicles that required large teams to modify, maintain, program, and considerable monetary support. Almost all current full-sized autonomous driving programs at companies and universities are direct descendants of, or are inspired by, the Urban Challenge. A lesson learned from the Urban Challenge about building a successful robot is that one should leverage existing technologies wherever possible. For example, an automotive company with decades of engineering and manufacturing expertise builds a much more reliable, robust platform than any team of researchers could. Also, time is freed to focus on the real unsolved problems in the space, mostly around the autonomy software.

More recently, automobile manufacturers, technology companies, automotive suppliers, and Universities have built full-sized autonomous vehicles that integrate drive-by-wire technology and the latest automotive grade sensors and computers into a streamlined, reliable package for engineers and researchers. Waymo \cite{waymo}, formerly the Google Autonomous Car project, is perhaps the most well-known fleet of autonomous vehicles on the road today with more than 100 vehicles operating in 4 cities and having logged more than 2 million autonomous miles in their 6 year history. All of these platforms range in price from \$300k to \$500k each and require large, dedicated development facilities, and carry major safety concerns.

Scaled platforms built from modified radio controlled cars are popular in the academic and hobby communities. Scaled platforms are typically 1 to 3 feet long and weigh between 5 and 50 pounds. They are orders of magnitude cheaper, can be built, maintained, and programmed by a small number of researchers, and they don't carry the safety and liability concerns of full sized vehicles, and have simple, standard interfaces to the sensors, computing, and actuators. The F1/10 Autonomous Racing Competition \cite{f110} allows teams to race against one another using a common 1:10 scale platform. All documentation for the robot is open source including designs, build instructions, and infrastructure software. The F1/10 platform is beneficial for researchers as they no longer have to spend time building one-off custom robots for their research and provides a benchmark system for comparing experimental results. However, 1:10 scale platforms are too small to exhibit realistic driving dynamics for testing control algorithms meant for full-sized vehicles, cannot carry an adequate sensor and computing payload for research algorithms, and currently available designs are not robust enough to survive repeatedly pushing the vehicle to the mechanical limits of the system.

\section{BEST RESPONSE MPPI}

Here we describe the control algorithm that we use for racing autonomously. The algorithm that we develop is a combination of a sampling based controller known as model predictive path integral control (MPPI), and the game theoretic notion of best response dynamics.

\subsection{Model Predictive Path Integral Control}

Model Predictive Path Integral (MPPI) is a sampling-based approach to model predictive control which has been successfully applied to aggressive autonomous driving using learned nonlinear dynamics~\cite{williams2016}. The underlying principle behind MPPI is to express the trajectory optimization problem as a probability matching problem. Let $V = \{\vv_0, \vv_1, \dots \vv_{T-1} \}$, and $U = \{\vu_0, \vu_1, \dots \vu_{T-1} \}$. We define $V$ as random control variables with mean $U$ such that: $V = U + \mathcal{E}$. Then, using an information theoretic lower bound, it is possible to show \cite{williams2016aggressive} that there exists an optimal distribution over controls, which is optimal in the sense that trajectories sampled from that distribution have a lower cost than any other control distribution. This distribution takes the form:
\begin{equation}
q^*(V) \propto \exp \left(-\frac{1}{\lambda}S(V)\right) p(V)
\end{equation}
where $S(V)$ is the state-dependent cost of a trajectory and $p(V)$ the probability density of $V$ from some prior distribution (e.g. zero mean Gaussian) which implicitly defines a control cost. The goal is to then minimize the KL-Divergence between the controlled an optimal distribution, which in turn leads to the following objective:
\begin{equation}
U^* = \argmin \left[ \frac{1}{2}\sum_{t=0}^{T-1} \left( \vu_t^\rT \Sigma^{-1} \vu_t - \int q^*(V) \vv_t \rd V \right) \right]
\end{equation}
which can be optimized by sampling trajectories from the (simulated) system dynamics, and computing a reward weighted average over sampled trajectories.

In a model predictive control setting, the algorithm starts with a planned control sequence $\left(u_0, u_1, \dots u_{T-1} \right) = U \in \mathbb{R}^{m \times T}$, and then samples a set of random control sequences $\left(\mathcal{E}_1, \mathcal{E}_2 \dots \mathcal{E}_K \right)$ , where each sequence consists of: $\mathcal{E}_k = \left(\epsilon_k^0, \dots \epsilon_k^{T-1}\right)$ and each $\epsilon_k^t \sim \mathcal{N}(u_t, \Sigma)$. Then the MPPI algorithm updates the control sequence as:
\begin{align}
\eta &= \sum_{k=1}^K \exp \left( -\frac{1}{\gamma} \left(S(V_k) + \lambda \sum_{t=0}^{T-1} u_t^\rT \Sigma^{-1} \epsilon_k^t \right) \right) \\
U &= \frac{1}{\eta}\sum_{k=1}^K \left [ \exp \left( -\frac{1}{\gamma} \left(S(V_k) + \lambda \sum_{t=0}^{T-1} u_t^\rT \Sigma^{-1} \epsilon_k^t \right) \right) \mathcal{E}_k \right]
\end{align}
The parameters $\gamma$ and $\lambda$ determine the selectiveness of the weighted average, and the importance of the control cost respectively. The key challenge for MPPI is sampling thousands of trajectories and evaluating them in real-time, which is achieved by parallelizing the sampling procedure using a GPU. Using a GPU, we can sample thousands of trajectories ($>1000$) in a fast real-time control loop (40 HZ). Algorithm \ref{Algorithm:mppi} describes the full MPPI algorithm.

\begin{algorithm}
  \SetKwInOut{Input}{Given}
  \Input{$\vF$: Transition Model\;
         $K$: Number of samples\;
         $T$: Number of timesteps\;
         $(\vu_0, \vu_1, ... \vu_{T-1})$: Initial control sequence\;
         $\Sigma, \phi, q, \lambda, \gamma$: Cost functions/parameters\;
         $\vu_{\min}, \vu_{\max}$: Actuator limits\;
         $\text{SGF}$: Savitsky-Galoy convolutional filter\;}        
  \While{task not completed}{
  $\vx_0 \leftarrow \text{GetStateEstimate()}$\;
  \For{$k \leftarrow 0$ \KwTo $K-1$}{
    $\vx \leftarrow \vx_0$\;
    Sample $\mathcal{E}^k = \left( \epsilon_0^k \dots \epsilon_{T-1}^k \right), ~\epsilon_t^k \in \mathcal{N}(0, \Sigma)$\;
    \For{$t \leftarrow 1$ \KwTo $T$}{
    $\vx_t \leftarrow \vF(\vx_{t-1}, \vu_{t-1} + \epsilon_{t-1}^k)$\;
      $S(\mathcal{E}^k) \pluseq  \vq(\vx_t) + \gamma \vu_{t-1}^\rT \Sigma^{-1} \epsilon_{t-1}^k$\;
    }
    $S(\mathcal{E}^k) \pluseq \phi(\vx_{T})$\;
  }
  $\beta \leftarrow \min_k[ S(\mathcal{E}^k) ]$\;
  $\eta \leftarrow \sum_{k=0}^{K-1} \exp\left( -\frac{1}{\lambda} (S(\mathcal{E}^k) - \beta) \right)$\;
  \For{$k \leftarrow 0$ \KwTo $K-1$}{
    $w(\mathcal{E}^k) \leftarrow \frac{1}{\eta}\exp\left( -\frac{1}{\lambda} (S(\mathcal{E}^k) - \beta) \right)$\;
  }
  \For{$t \leftarrow 0$ \KwTo $T-1$}{
  $U \leftarrow U + \text{SGF}*\left(\sum_{k=1}^K w(\mathcal{E}^k) \mathcal{E}\right)$\;
  }

  $\text{SendToActuators}(\vu_0)$\;

  \For{$t \leftarrow 1$ \KwTo $T-1$}{ 
      $\vu_{t-1} \leftarrow \vu_t$\;  
  }
  $\vu_{T-1} \leftarrow \text{Intialize}(\vu_{T-1})$\;
  }
  \label{Algorithm:mppi}
  \caption{MPPI}
\end{algorithm}

\subsection{Elementary Game Theory}

The game theoretic concept of an N-player ``game" consists of a set of possible actions (or strategies) for each player $\calU = \left( U_1, \dots U_N \right)$, and an objective function (which in our setting is to be minimized) for each player $\calJ = \left( J_1, \dots J_N \right)$. We can then define a game as the tuple:
\begin{equation}
\left(\calU, \calJ \right)
\end{equation}
Given the assumption that each player is acting rationally (i.e. attempting to minimize cost), we need an appropriate solution concept. In the case of a one player  ``game" it is simply an optimization problem. However, in the multi-player case each players objective potentially depends on the actions of every other player, so it is not possible to find a solution set $\mathcal{U} = \left\{ U_1 \dots U_N \right\}$ such that:
\begin{equation}
\forall i \in \{1, \dots N\}, ~ U_i = \argmin_{W \in \calA}\left[ J_i \left( W \right) \right] 
\end{equation}
So we need to develop alternative notions of solutions. In our setting we assume that players do not communicate strategy or explicitly cooperate (although in some cases there may be an illusion of cooperation if the cost functions have mutually beneficial terms such avoiding collisions). The solution concept most appropriate in this case is the Nash Equilibrium.
\begin{definition}
A set of strategies $\left(U_1^*, \dots U_N^* \right)$ are said to be in a Nash Equilibrium if
\begin{equation*}
\forall i \in \{1, \dots N\},~ U_i^* = \argmin_{U_i}\left[ J\left(U_i, U_{-i} \right)\right]
\end{equation*}
where $U_{-i} = \left\{ U_1 \dots U_N \right \} \setminus \left\{ U_i \right \}$. That is, each players strategy is optimal given that the strategies of all the other players is fixed.
\end{definition}
If a game is in Nash equilibrium it means that no player can improve their objective by unilaterally changing their strategy. For general continuous games, finding a Nash Equilibrium is very difficult, which means that it is a poor fit for real-time control. So, instead of focusing on finding a Nash Equilibrium we will pursue a different model of agent behavior which, in the best case, converges to a Nash equilibrium, but in the worst case may cycle between solutions which results in indecision.

\subsection{Best Response Dynamics}

The fundamental object that we consider for modeling agent behavior is the best-response set. Consider agent $i$, and a set of opponent strategies $U_{-i}$, the set of best responses for agent $i$ to the opponent strategies is:
\begin{equation}
R_i = \left\{U_i ~\big|~  J(U_i, U_{-i}) = \min_{W_i} J(W_i, W_{-i}) \right\}
\end{equation}
In other words the best-response set is the set of strategies which minimize the agents cost given the opponents strategies as fixed. The \emph{best response dynamics} for a game is the dynamical system obtained by iteratively choosing a best response strategy for each agents based on the fixed behaviors of the other agents. Formally let $f_i(U_{-i})$ be a function which takes in the current set of strategies for the other players and returns a strategy from the best response set. Then the best response dynamics are:
\begin{align*}
U_1^{k+1} &= f_1(U_2^k, U_3^k \dots U_N^k) \\
U_2^{k+1} &= f_2(U_1^k, U_3^k \dots U_N^k) \\
&\vdots \\
U_N^{k+1} &= f_N(U_1^k, U_2^k \dots U_{N-1}^k)
\end{align*}
This is an intuitively appealing model of strategic thinking for real-time control, based on the best guess of their opponents strategies each player picks a strategy to minimize their own cost while considering their opponents strategies as fixed. Finding this update is a standard (one-player) optimization problem, and the updates can all be performed in parallel which means that it is potentially computationally efficient. The best response dynamics also have a strong connection to the solution concept of Nash Equilibrium described by the following theorem in that a set of strategies $(U_1, U_2 \dots U_N)$ is a Nash Equilibrium if and only if it is an equilibrium point of the best response dynamics.

So, if the best response dynamics converge to a point it is necessarily to a Nash Equilibrium solution. In general, the best response dynamics need not converge, they can converge to a limit cycle (think of rock-paper-scissors for example). However, this may not be a pressing problem in the context of real-time control. The system is always changing, so even if we're become momentarily stuck in a limit cycle there's always the chance to jump out of it as soon as the control system evolves. Additionally, the goal is to model the behavior of the other agent which the best-response dynamics may still do in this regard, even without converging to a Nash equilibrium.

\subsection{BR-MPPI Algorithm}

We can combine the best response model of opponent behavior with the model predictive path integral control algorithm by including the predicted opponents trajectory (without noise) into the cost of the other vehicle. This means that the state-dependent cost of an input control sequences for agent $i$ becomes:
\begin{equation}
S(V_i, U_{-i})
\end{equation}
And from the information theoretic perspective of minimizing the KL-Divergence between the controlled and optimal distribution, we obtain the following objective for agent $i$:
\begin{equation}
\min \left[ \frac{1}{2}\sum_{t=0}^{T-1} \left( (\vu_t^i)^\rT \Sigma^{-1} \vu_t^i - \int q^*(V_i, U_{-i}) \vv_t^i \rd V_i \right) \right]
\end{equation}
where the optimal distribution $q(V_i, U_{-i})$ is tied to the actions of the other agents through the state-dependent cost $S(V_i, U_{-i})$. The best response for each agent is then:
\begin{equation}
w_k^i = \exp \left( -\frac{1}{\gamma} \left(S(V_k, U_{-i}) + \lambda \sum_{t=0}^{T-1} (u_t^i)^\rT \Sigma^{-1} \epsilon_{k,t}^i \right) \right)
\end{equation}
\begin{equation}
U_i = \frac{\sum_{k=1}^K w_k^i \mathcal{E}_k }{\sum_{k=1}^K w_k^i}
\end{equation}
For each nearby agent, the best response MPPI (BR-MPPI) algorithm maintains a nominal control policy which is updated by sampling many possible trajectories for each agent, and computing a cost-weighted average over the sampled trajectories where the cost evaluation takes place by considering the nominal policies of the other agents as fixed. In this paper, we consider open-loop control sequences as the control policy, and only consider a single adversarial opponent.

\begin{algorithm}
  \SetKwInOut{Input}{Given}
  \Input{$\vF_1, \dots \vF_n$: Transition models \;
         $S_1, \dots S_n$: Agent Costs;}        
  \While{task not completed}{
	$U_1 \leftarrow MPPI(\vF_1, S_1, \mathcal{X}_{-1})$\;
  	\For{$t \leftarrow 0$ \KwTo $T-1$}{
  	$\mathcal{X}_1^t \leftarrow \vx $\;
    $\vx \leftarrow \vF_1(\vx, \vu_t^1)$\;
  	}
    $\cdots$\\
  	\For{$t \leftarrow 0$ \KwTo $T-1$}{
  	$\mathcal{X}_N^t \leftarrow \vx $\;
    $\vx \leftarrow \vF_N(\vx, \vu_t^N)$\;
  	}   
    $U_N \leftarrow MPPI(\vF_N, S_N, \mathcal{X}_{-N})$\;
  }
  \label{Algorithm:BR-MPPI}
  \caption{BR-MPPI}
\end{algorithm}

\section{EXPERIMENTAL SETUP}

The AutoRally robot is a robust all-electric autonomous testbed that is 1:5 the size of passenger vehicles. The robot has an onboard Mini-ITX form factor computer housed in a rugged enclosure. The computer configuration is shown in Table~\ref{table:computeBoxDet}. The sensor package on AutoRally includes 2 forward facing cameras, a Lord Microstrain 3DM-GX4-25 IMU, an RTK corrected GPS receiver, and hall effect wheel speed sensors. The entire robot weighs approximately 22 kg and measures 0.9 meters from front to back. This platform allows for the self-contained testing of all algorithms, with no reliance on external position systems or computation beyond a GPS receiver.

\begin{table}
   \centering
   \caption{AutoRally compute box computing and power components.}
  \begin{tabular}{l | c }
    \textbf{Component} & \textbf{Detail}\\
    \hline
    Motherboard & Asus Z170I Pro Gaming, Mini-ITX \\
    CPU & Intel i7-6700, 3.4 GHz quad-core 65 W \\
    RAM & 32 GB DDR4, 2133 MHz\\
    GPU & Nvidia GTX-750ti SC, 640 cores, 2 GB, 1176 MHz\\
    SSD storage & 512 GB M.2 and 1 TB SATA3\\
    Wireless & 802.11ac WiFi, 900 MHz XBee, and 2.4 GHz RC \\
    Power supply & Mini-Box M4-ATX, 250 W \\
  	Battery & 22.2 V, 11 Ah LiPo, 244 Wh \\
  \end{tabular}
  \label{table:computeBoxDet}
\end{table}

\begin{figure}
\centering
\includegraphics[width = .95\columnwidth]{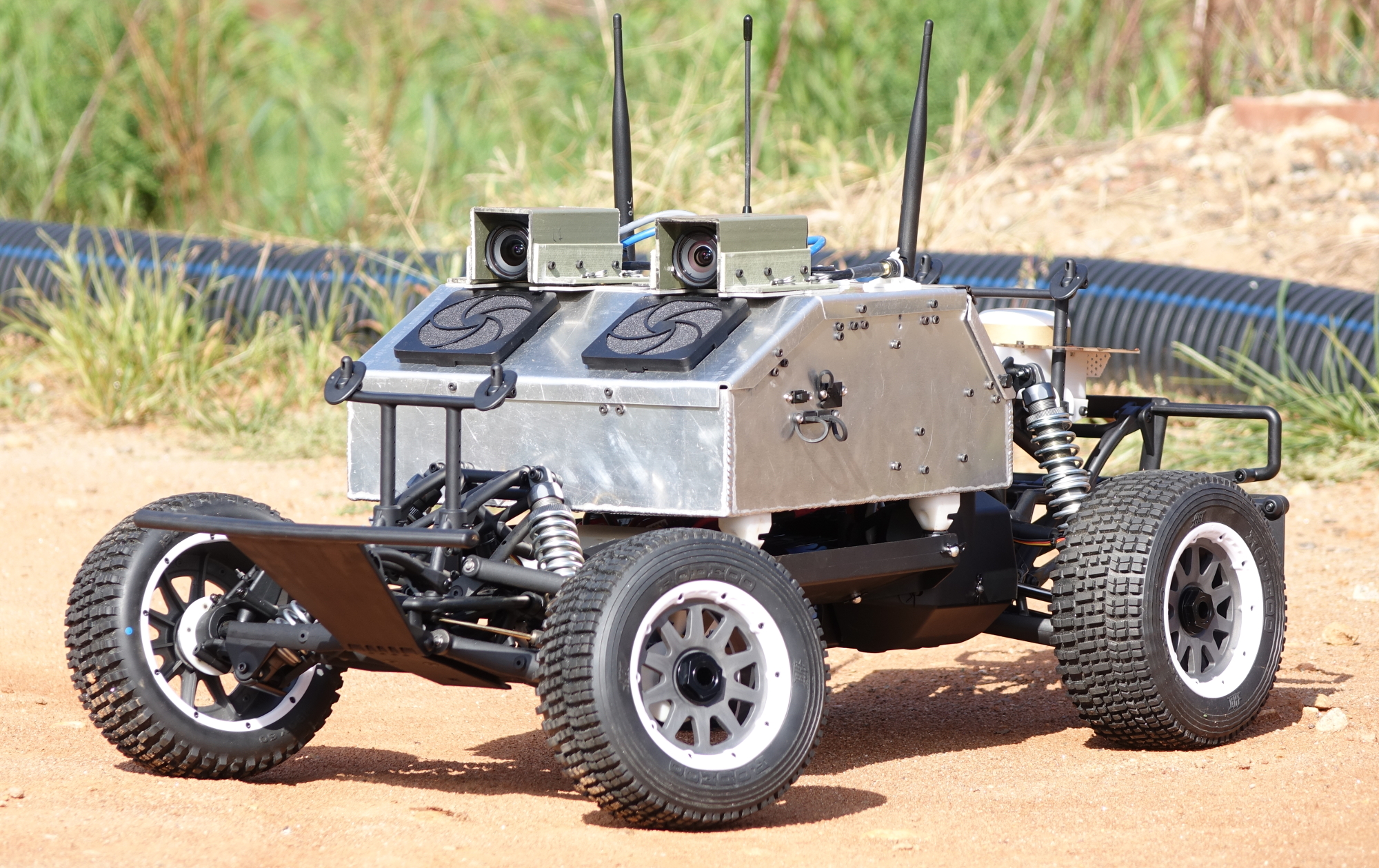}
\caption{AutoRally platform with protective body removed.}
\label{Figure:autorally}
\end{figure}

An accurate state estimate for each vehicle is computed using GPS and IMU measurement, along with the software packages GTSAM and iSAM2\cite{isam2}.

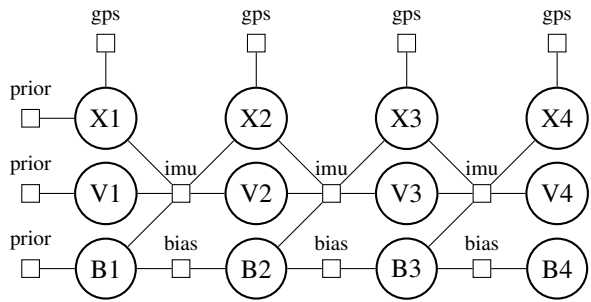
\begin{figure}
\centering
\begin{tikzpicture}
 \tikzstyle{state}=[circle,thick,draw=black] (1)
 \tikzstyle{factor}=[rectangle,draw=black] (1,1)
 \tikzstyle{factoredge}=[draw]

 \node [state] (x4) {X4};
 \node [state] (v4) [below of=x4] {V4};
 \node [factor] (imu3) [left of=v4, label=above:\footnotesize imu] {};
 \node [state] (v3) [left of=imu3] {V3};
 \node [factor] (imu2) [left of=v3, label=above:\footnotesize imu] {};
 \node [state] (v2) [left of=imu2] {V2};
 \node [factor] (imu1) [left of=v2, label=above:\footnotesize imu] {};
 \node [state] (v1) [left of=imu1] {V1};
 \node [state] (x3) [above of=v3] {X3};
 \node [state] (x2) [above of=v2] {X2};
 \node [state] (x1) [above of=v1] {X1};

 \node [state] (b4) [below of=v4] {B4};
 \node [state] (b3) [below of=v3] {B3};
 \node [state] (b2) [below of=v2] {B2};
 \node [state] (b1) [below of=v1] {B1};

 \node [factor] (cb1) [left of=b2, label=above:\footnotesize bias] {};
 \node [factor] (cb2) [left of=b3, label=above:\footnotesize bias] {};
 \node [factor] (cb3) [left of=b4, label=above:\footnotesize bias] {};

 \node [factor] (p1) [left of=x1, label=above:\footnotesize prior] {};
 \node [factor] (p2) [left of=v1, label=above:\footnotesize prior] {};
 \node [factor] (p3) [left of=b1, label=above:\footnotesize prior] {};

 \node [factor] (gps1) [above of=x1, label=above:\footnotesize gps] {};
 \node [factor] (gps2) [above of=x2, label=above:\footnotesize gps] {};
 \node [factor] (gps3) [above of=x3, label=above:\footnotesize gps] {};
 \node [factor] (gps4) [above of=x4, label=above:\footnotesize gps] {};

 \foreach \from/\to in {cb1/b1,cb1/b2,cb2/b2,cb2/b3,cb3/b3,cb3/b4}
   \draw (\from) -- (\to);
 \foreach \from/\to in {p1/x1,p2/v1,p3/b1}
   \draw (\from) -- (\to);
 \foreach \from/\to in {gps1/x1,gps2/x2,gps3/x3,gps4/x4}
   \draw (\from) -- (\to);
 \foreach \from/\to in {imu1/x1,imu1/x2,imu1/v1,imu1/v2,imu1/b1,imu2/x2,imu2/x3,imu2/v2,imu2/v3,imu2/b2,imu3/x3,imu3/x4,imu3/v3,imu3/v4,imu3/b3}
   \draw (\from) -- (\to);

\end{tikzpicture}
\caption{Factor Graph structure used for GPS/IMU fusion, circles represent states and squares represent factors.}
\label{Figure:FactorGraph}
\end{figure}

\subsection{Dynamics Models}

One of the key components of the BR-MPPI algorithm is using forward dynamics models of the vehicles to sample trajectories. This is computationally demanding, since each dynamics model must be evaluated millions of times per second, and the non-linear dynamics of the vehicle are not easy to model analytically. To alleviate this difficulty, we use two dynamics models: an expensive model for the vehicle that is actually being controlled, and a cheap model for the other agents. The idea is that, since the controls for the other vehicles aren't actually applied, we can use a cheaper less accurate model for the other vehicles as long as it captures the essential constraints in the dynamics. For a cheap model we use a linear basis function model, and for an expensive model we use a multi-layer neural network. 

\subsubsection{Basis Function Model}
The basis function model supposes that the dynamics take the form:
\begin{equation}
\vf(\vx_d) = \Theta^\rT \phi(\vx_d)
\end{equation}
where $\Theta \in \Rb^{b \times 4}$ and $\phi(\vx) = (\phi_1(\vx), \phi_2(\vx), \dots \phi_b(\vx))^\rT \in \Rb^b$ is a matrix of coefficients and a vector of non-linear basis functions respectively. Given this model form, there are two challenges: determining an appropriate set of basis functions, and computing the coefficient matrix $\Theta$. 

For determining the appropriate set of basis function we analyzed the non-linear bicycle model of vehicle dynamics from \cite{hindiyeh2013dynamics} and extracted out all of the non-linear functions that appeared in the algebraic equations. This led to a set of 21 basis functions, and then, based on trial and error, we added 4 more to account for the roll dynamics and the non-linear throttle calibration. Given a set of basis functions and some data collected from the system, determining the coefficient matrix $\Theta$ is an unconstrained linear regression problem which can be solved in closed form.

\subsubsection{Neural Network Model}
The second model that we trained  to approximate the dynamics function, $\vf(\vx_d)$, was a multi-layer neural network model. We use a two-layer fully connected model with hyperbolic tangent non-linearities. The neural network model is trained using a 30 minute system identification dataset.

\subsection{Vehicle to Vehicle Communication}

Each AutoRally robot operates as a self-contained system, but is equipped with multiple wireless radios for communications. We bypass the standard detection and tracking problem required for multi-vehicle racing by extending the functionality of the XBee radio on each robot to include vehicle to vehicle pose communication. The 900 MHz XBee provides a high-reliability, low bandwidth mesh network that all vehicles at the test site connect to.
\begin{figure}[b]
\centering
\includegraphics[width = .95\columnwidth]{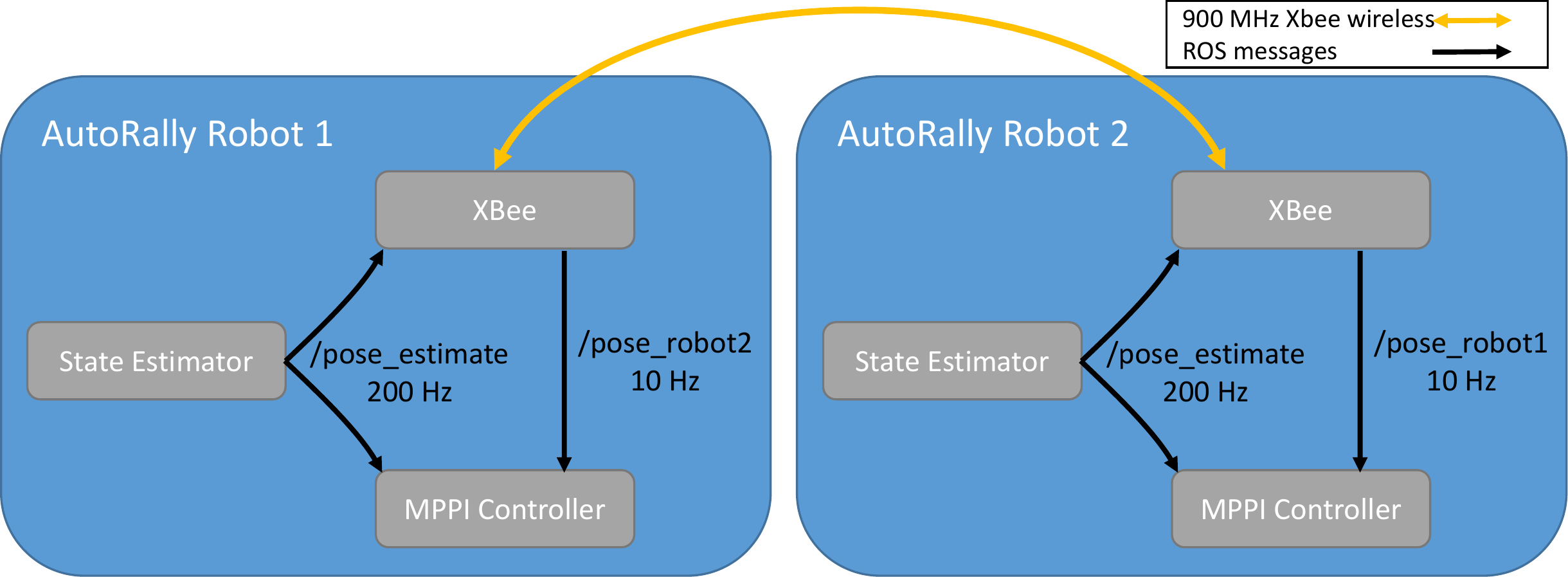}
\caption{Vehicle to vehicle system for broadcasting state estimates at 10 Hz from one robot to all other robots within communications range.}
\label{Figure:wirelessComms}
\end{figure}

Each robot runs a standalone state estimator that fuses IMU and GPS information to produce a very accurate state estimate at 200 Hz. The high rates are necessary for high speed, real time control, but would quickly saturate the XBee network as the number of vehicles within communications range increases. For that reason, we implemented a configurable rate, currently set at 10Hz, to down sample pose estimates and then transmit them over the XBee network. Figure~\ref{Figure:wirelessComms} shows the wireless pose communication system for two vehicles and the routing of signals within each robot between the state estimator, XBee software, and MPPI controller. In practice, this communication system works well with two vehicles at the test site, and the XBee radios are much more reliable than using the WiFi network.

\subsection{Cost Function Design}

The base cost function for the BR-MPPI controller consists of a term for staying on the track, a cost for going a certain speed, a control cost, and some additional cost terms which helps stabilize the vehicle. This base portion of the cost functions can be described as:
\begin{equation}
c(x) = w \cdot \left(C_M(p_x, p_y), (v_x - v_x^d)^2, 0.9^t I,  \left(\frac{v_y}{v_x}\right)^2 \right)
\end{equation}
where $C_M(p_x, p_y)$ is a function which returns a value of 1 if the vehicle is off the track, a value of $0$ is it is in the center of the track, and smoothly interpolates between zero and 1 for values on the track but not directly in the center. The second term $(v_x - v_x^d)^2$ forces the vehicle to go a certain speed, the third term is an indicator variable denoting if the vehicle has crashed or not, and the last term is a quadratic penalty on the side-slip angle of the vehicle.

In addition to this base cost function, for autonomous racing, we added a new term into the cost function which penalizes collisions and provides rewards and penalties for performing passing maneuvers. Let $\vz_{t-1}$ and $\vz_t$ be the projected location of the opposing vehicle in the body frame coordinate system of the robot. If $\vz$ is in the right-half plane then it means that the opposing robot is in front of the opposing vehicle, and if $\vz$ is in the left-half plane it means that the opposing robot is behind the vehicle. The robot gets a negative cost (set at -5000) if it goes from trailing to leading the opposing vehicle, and a positive penalty (set at 5000) if it goes from leading to trailing. Additionally, if the vehicle enters collision \emph{when it is the trail vehicle} then it receives a penalty (set at 10000). We found that if we enforced this penalty at all times then the vehicle would be too timid, and often move out of the way of the pursuing vehicle which is not desirable in a racing context.

\subsection{Opponent Vehicle}

As an opponent for the vehicle we had a human pilot control the opposing vehicle. The human pilot is capable of driving at similar top speeds to the base MPPI algorithm, so racing between the two is competitive. 

\section{RESULTS}

\begin{figure}[h!]
\centering
\includegraphics[width=.95\columnwidth]{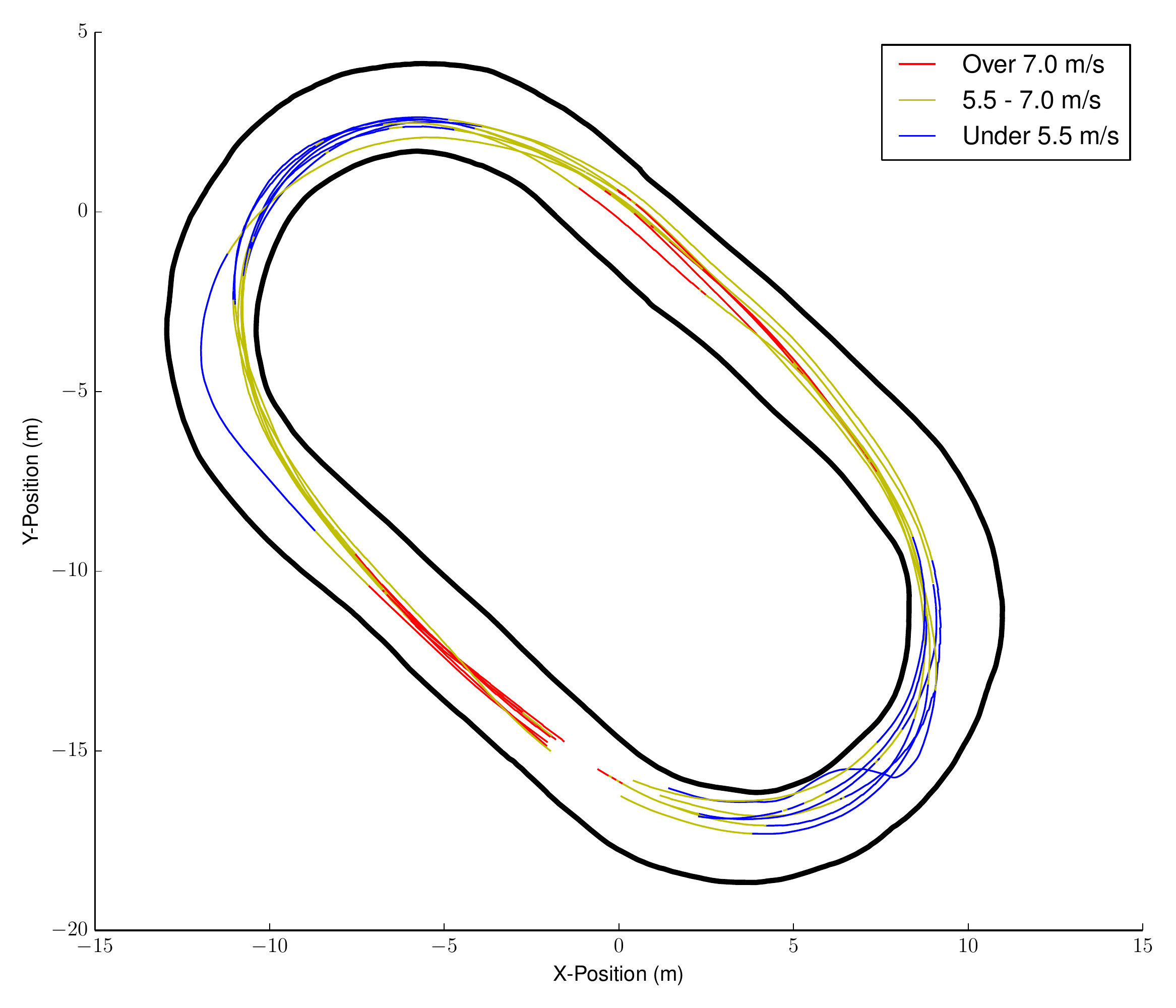}

\includegraphics[width=.95\columnwidth]{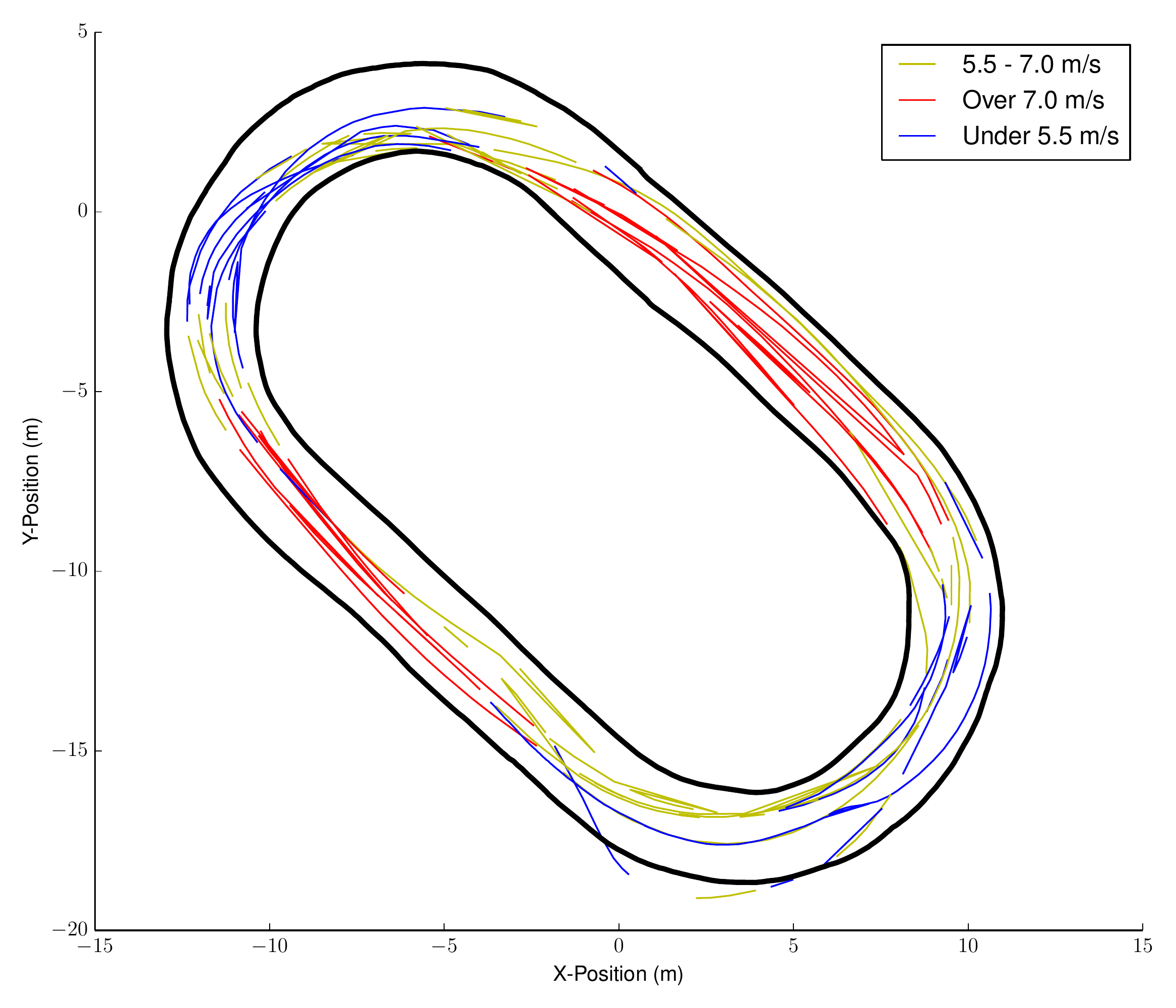}
\caption{Trajectory traces of the two vehicles racing against each other. Left: BR-MPPI controlled vehicle. Right: Human controlled vehicle for the same trial. The two figures are synced in time.}
\label{Fig:TrajTraces}
\end{figure}

The BR-MPPI algorithm did show the capability of driving around the track successfully while trailing or leading the opponent vehicle. Figure \ref{Fig:TrajTraces} shows a mostly successful trial run where the MPPI controlled vehicle races against a human controlled vehicle. The BR-MPPI algorithm carries more speed into corners and is more consistent than the human driver, however the human attains a higher top speed along the straights which allows the human controlled vehicle to keep up with the BR-MPPI algorithm. 

Overall, the success rate at avoiding collisions while outperforming the human controlled vehicle is not very high. However, the algorithm shows some promising signs, namely reactiveness and the ability to perform intelligent passing maneuvers, that suggest small changes to the cost function or optimization procedure could result in a high rate of success.

\subsubsection{Reactiveness to Opponent}

\begin{figure}[h!]
\centering
\includegraphics[width=.3\columnwidth]{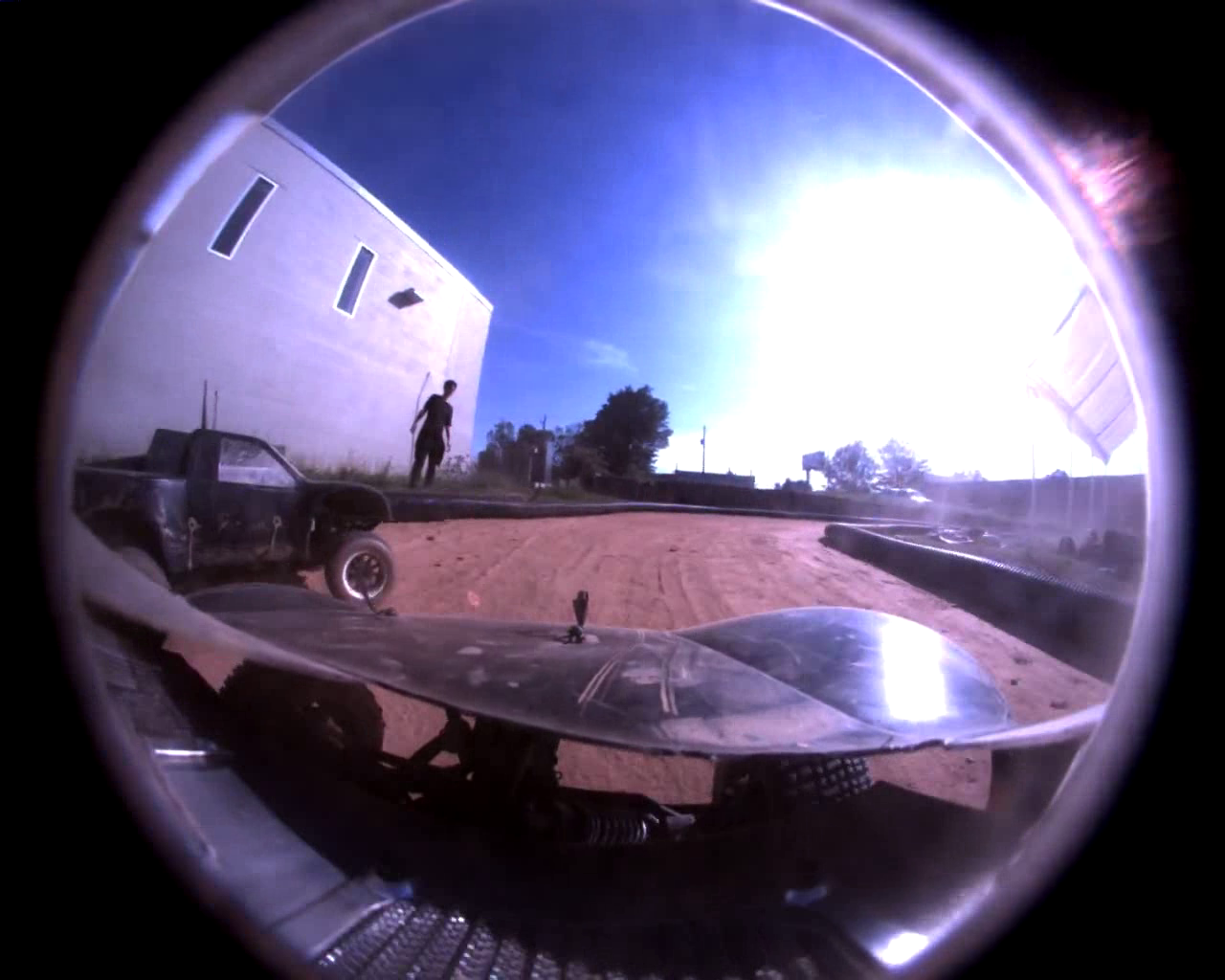}
\includegraphics[width=.3\columnwidth]{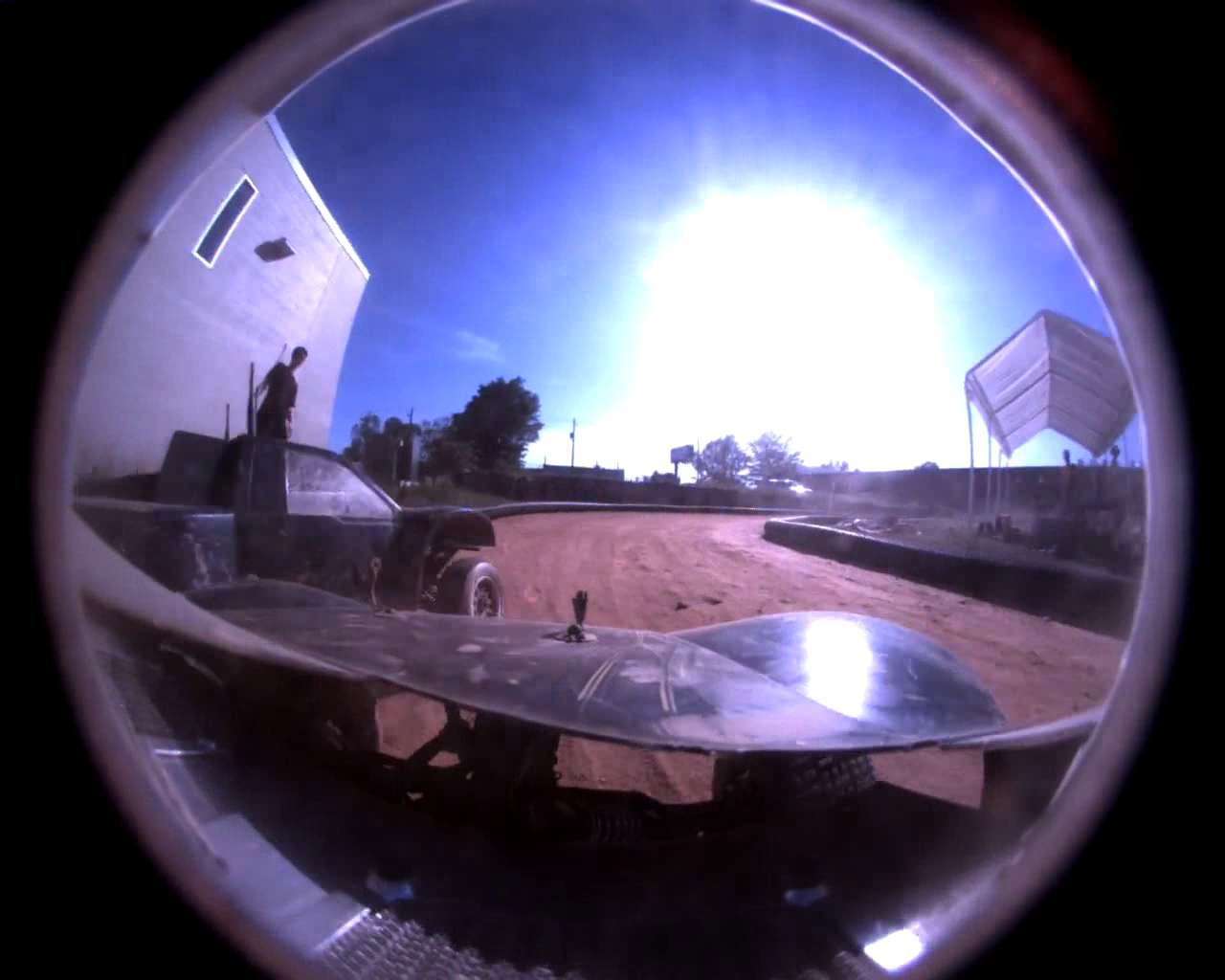}
\includegraphics[width=.3\columnwidth]{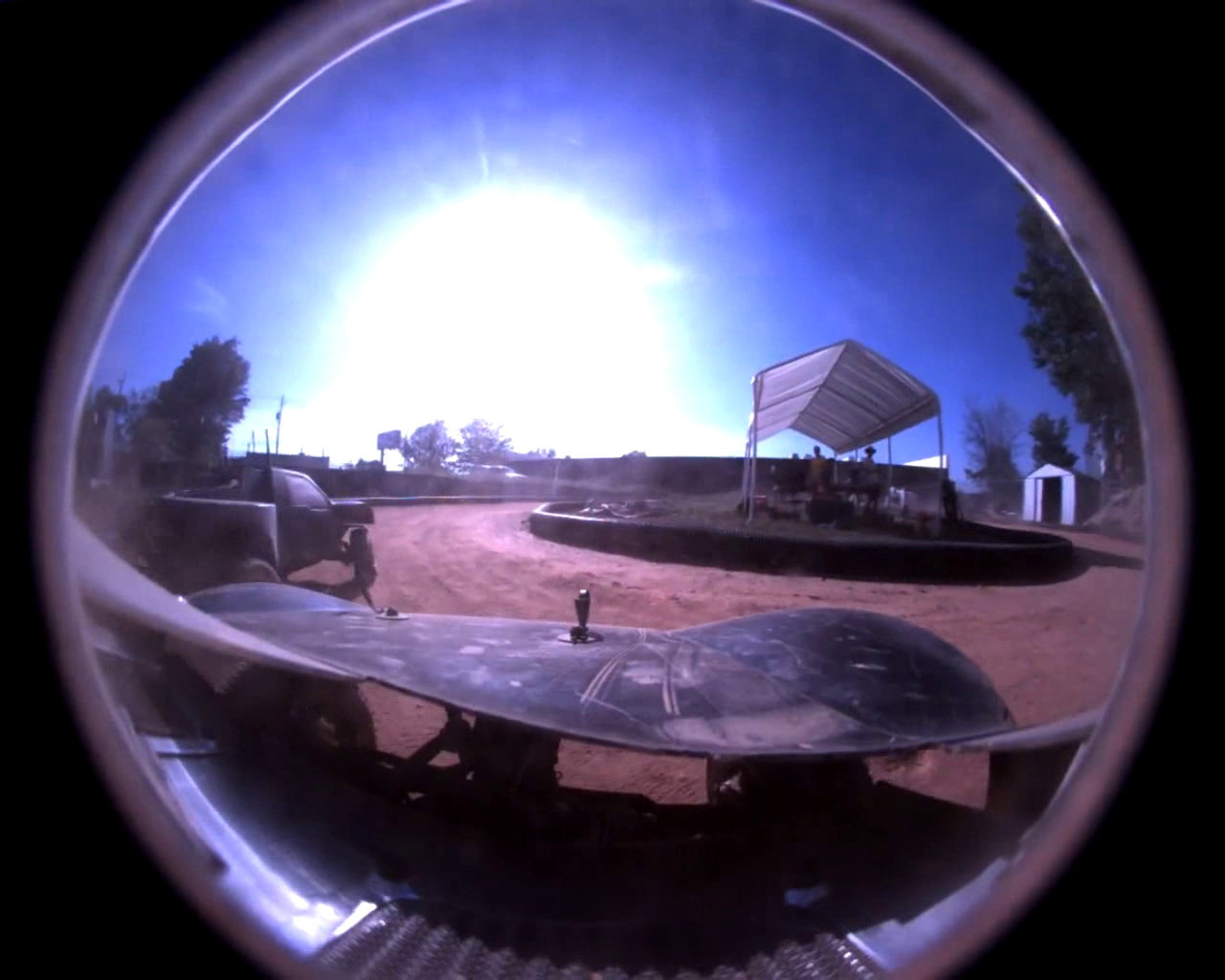}
\caption{Collision avoidance sequence. The opponent vehicle slides wide into the turn and initially is predicted to slide off the track. However, it recovers and forces the autonomous vehicle to avoid a collision.}
\label{Fig:CollisionAvoidance}
\end{figure}

\begin{figure}[h!]
\centering
\includegraphics[width=.95\columnwidth]{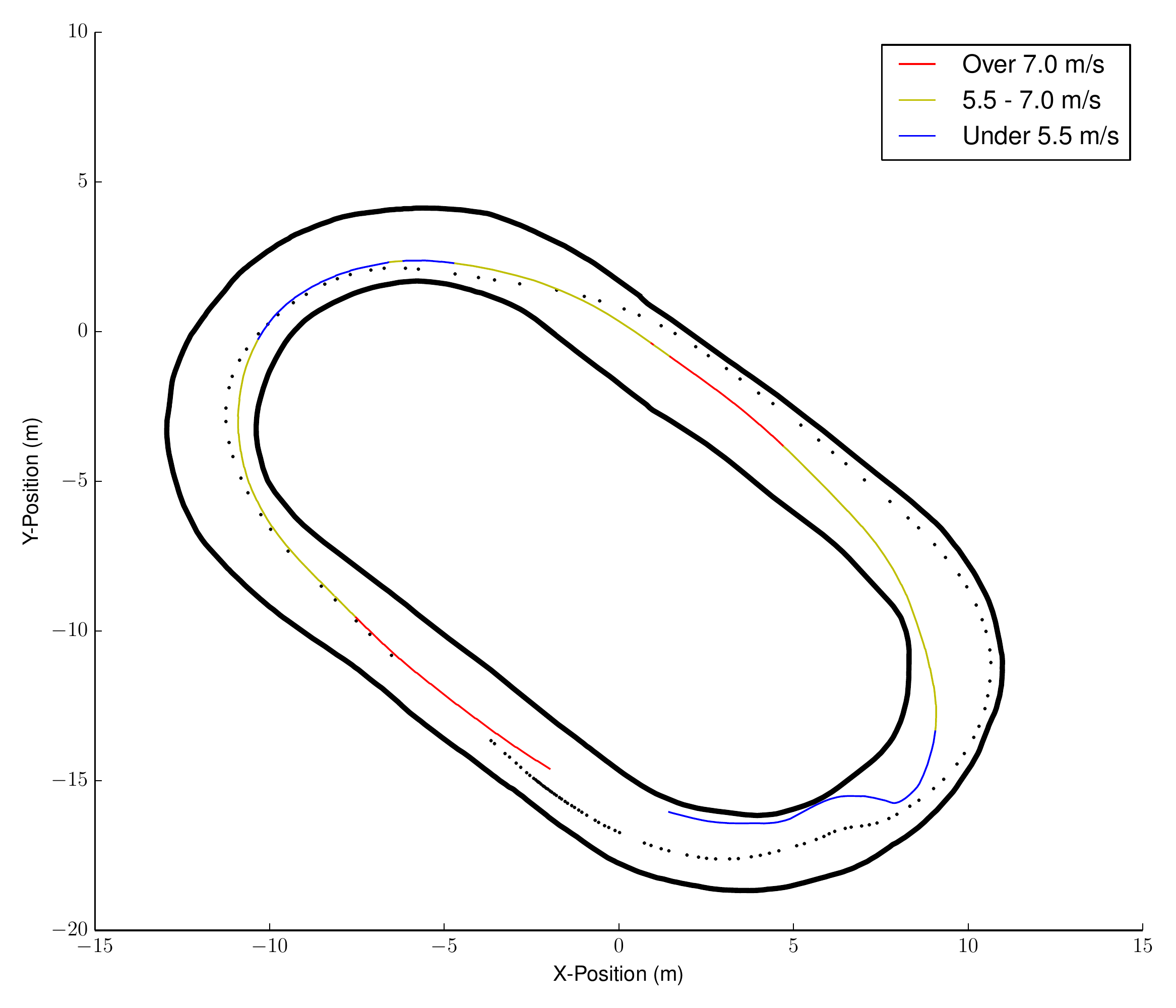}
\caption{Trajectory traces of BR-MPPI controlled vehicle (straight line) and the human controlled vehicle (dotted) when narrowly avoiding a collision}
\label{Fig:Maneuvers}
\end{figure}

The BR-MPPI algorithm demonstrated the ability to react very quickly to opponent actions in order to avoid collisions. Figures \ref{Fig:CollisionAvoidance} and \ref{Fig:Maneuvers} demonstrates this capability. In this example the opponent vehicle swings wide going into the turn, and the BR-MPPI algorithm initially predicts that the opponent vehicle will slide off the track. However, the opponent vehicle ends up recovering and cutting inward in front of the BR-MPPI vehicle's planned trajectory. As a result, the BR-MPPI vehicle has to quickly respond and does so by hitting full brake and turning to avoid collision. 

\subsection{Passing Maneuvers}
\begin{figure}[!h]
\centering
\includegraphics[width=.95\columnwidth]{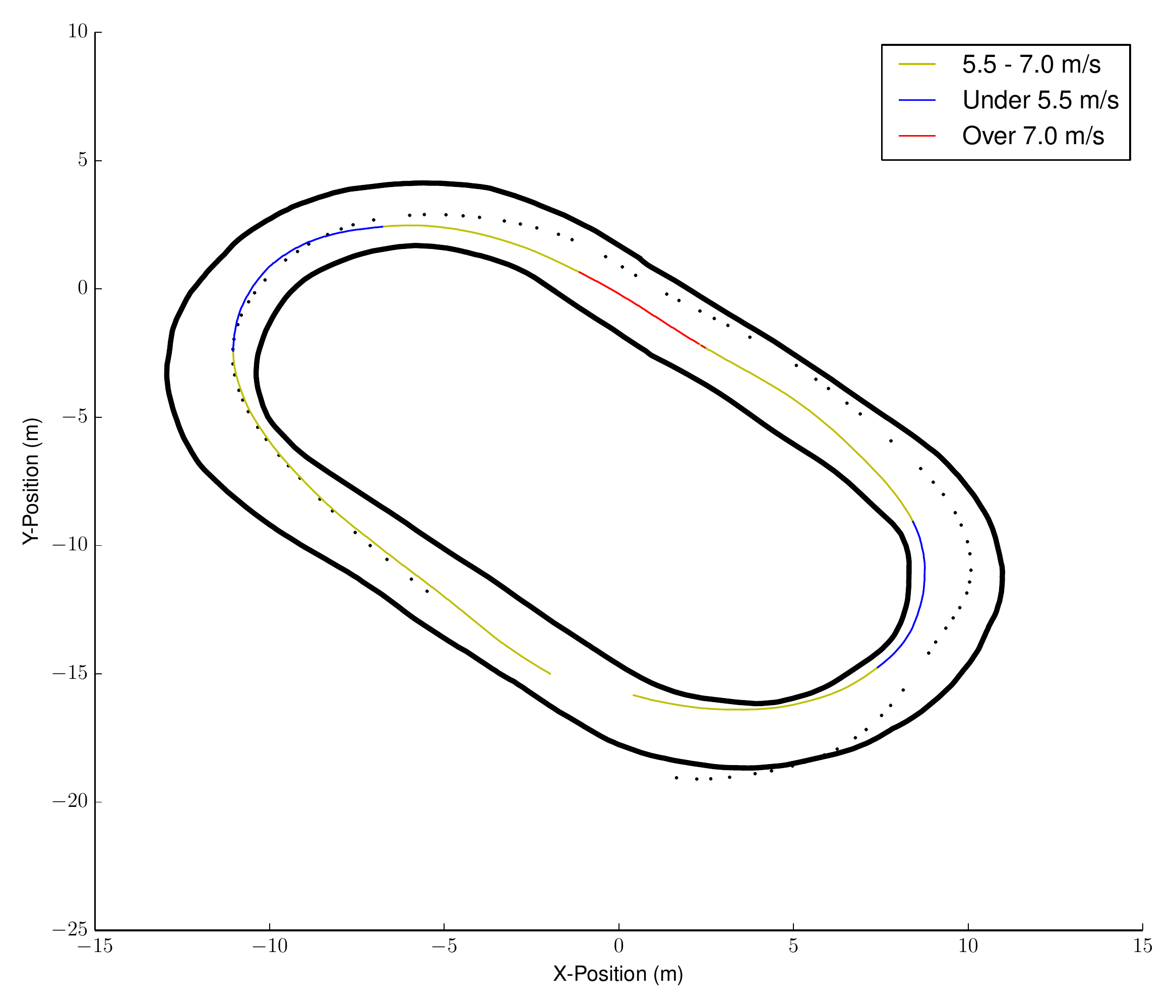}
\caption{Trajectory traces of BR-MPPI controlled vehicle (straight line) and the human controlled vehicle (dotted) when performing a passing maneuver while going around a corner.}
\label{Fig:Passing}
\end{figure}

The most encouraging aspect of the BR-MPPI algorithms behavior was its ability to perform intelligent passing maneuvers, this is especially impressive given the very limited width of the test track. Figure \ref{Fig:Passing} demonstrates this ability, where the passing maneuver is performed on the portion of the track in the lower right of the figure. The vehicle and its opponent enter the turn nearly side by side, and the BR-MPPI algorithm plans a trajectory which tightly hugs the inside corner. Notice how the cornering maneuver performed in Fig. \ref{Fig:Passing} is much different than the average line taken by the BR-MPPI algorithm (Fig. \ref{Fig:TrajTraces}). This is evidence that the BR-MPPI controller is capable of accurately planning and executing intelligent racing behaviors.

\section{CONCLUSION}
In this paper, we extended previous work on sampling-based model predictive control to the multi-vehicle racing domain with the Best-Response Model Predictive Path Integral Control (BR-MPPI) algorithm. To bypass the perception problem in the racing scenario, we implemented vehicle-to vehicle communications on the AutoRally platform so that each vehicle transmits its state estimate and receives the state estimate from other robots over the included XBee radio. We collected real world experimental results using two AutoRally platforms at the Georgia Tech Autonomous Racing Facility with one driving autonomously using the BR-MPPI algorithm and the other operated by a skilled human. Results show that BR-MPPI is competitive with the human operator, and drives a much more consistent path than the human. The autonomous robot and human operator were unable to pass each other at the highest speeds unless the lead vehicle made a mistake. While the results are promising, there is room to improve BR-MPPI driving performance with parameter tuning and improved dynamics models.

\bibliographystyle{IEEEtran}
\bibliography{references}

\begin{thebibliography}{10}
\providecommand{\url}[1]{#1}
\csname url@rmstyle\endcsname
\providecommand{\newblock}{\relax}
\providecommand{\bibinfo}[2]{#2}
\providecommand\BIBentrySTDinterwordspacing{\spaceskip=0pt\relax}
\providecommand\BIBentryALTinterwordstretchfactor{4}
\providecommand\BIBentryALTinterwordspacing{\spaceskip=\fontdimen2\font plus
\BIBentryALTinterwordstretchfactor\fontdimen3\font minus
  \fontdimen4\font\relax}
\providecommand\BIBforeignlanguage[2]{{%
\expandafter\ifx\csname l@#1\endcsname\relax
\typeout{** WARNING: IEEEtran.bst: No hyphenation pattern has been}%
\typeout{** loaded for the language `#1'. Using the pattern for}%
\typeout{** the default language instead.}%
\else
\language=\csname l@#1\endcsname
\fi
#2}}

\bibitem{isaacs1999differential}
\BIBentryALTinterwordspacing
R.~Isaacs, \emph{Differential Games: A Mathematical Theory with Applications to
  Warfare and Pursuit, Control and Optimization}, ser. Dover books on
  mathematics.\hskip 1em plus 0.5em minus 0.4em\relax Dover Publications, 1999.
  [Online]. Available: \url{https://books.google.com/books?id=XIxmMyIQgm0C}
\BIBentrySTDinterwordspacing

\bibitem{morimoto2003minimax}
J.~Morimoto, G.~Zeglin, and C.~G. Atkeson, ``Minimax differential dynamic
  programming: Application to a biped walking robot,'' in \emph{Intelligent
  Robots and Systems, 2003.(IROS 2003). Proceedings. 2003 IEEE/RSJ
  International Conference on}, vol.~2.\hskip 1em plus 0.5em minus 0.4em\relax
  IEEE, 2003, pp. 1927--1932.

\bibitem{walrand2011harbor}
J.~Walrand, E.~Polak, and H.~Chung, ``Harbor attack: A pursuit-evasion game,''
  in \emph{Communication, Control, and Computing (Allerton), 2011 49th Annual
  Allerton Conference on}.\hskip 1em plus 0.5em minus 0.4em\relax IEEE, 2011,
  pp. 1584--1591.

\bibitem{gill2005snopt}
P.~E. Gill, W.~Murray, and M.~A. Saunders, ``Snopt: An sqp algorithm for
  large-scale constrained optimization,'' \emph{SIAM review}, vol.~47, no.~1,
  pp. 99--131, 2005.

\bibitem{pan2012pursuit}
S.~Pan, H.~Huang, J.~Ding, W.~Zhang, C.~J. Tomlin, \emph{et~al.}, ``Pursuit,
  evasion and defense in the plane,'' in \emph{American Control Conference
  (ACC), 2012}.\hskip 1em plus 0.5em minus 0.4em\relax IEEE, 2012, pp.
  4167--4173.

\bibitem{sadigh2016planning}
D.~Sadigh, S.~Sastry, S.~A. Seshia, and A.~D. Dragan, ``Planning for autonomous
  cars that leverage effects on human actions.'' in \emph{Robotics: Science and
  Systems}, 2016.

\bibitem{buehler2009urban}
M.~Buehler, K.~Iagnemma, and e.~Singh, Sanjiv, \emph{The DARPA urban challenge:
  autonomous vehicles in city traffic}, ser. Springer Tracts in Advanced
  Robotics.\hskip 1em plus 0.5em minus 0.4em\relax Springer, 2009.

\bibitem{waymo}
``Waymo (formerly google autonomous car),'' \url{https://waymo.com/}, accessed:
  2017-02-10.

\bibitem{f110}
``F1/10 autonomous racing competition,'' \url{http://f1tenth.org/}, accessed:
  2017-02-10.

\bibitem{williams2016}
G.~Williams, P.~Drews, B.~Goldfain, J.~M. Rehg, and E.~A. Theodorou,
  ``Aggressive driving with model predictive path integral control,'' in
  \emph{IEEE International Conference on Robotics and Automation (ICRA)}, May
  2016, pp. 1433--1440.

\bibitem{williams2016aggressive}
------, ``Aggressive driving with model predictive path integral control,'' in
  \emph{Robotics and Automation (ICRA), 2016 IEEE International Conference
  on}.\hskip 1em plus 0.5em minus 0.4em\relax IEEE, 2016, pp. 1433--1440.

\bibitem{isam2}
M.~Kaess, H.~Johannsson, R.~Roberts, V.~Ila, J.~J. Leonard, and F.~Dellaert,
  ``\BIBforeignlanguage{{English}}{{iSAM2: Incremental smoothing and mapping
  using the Bayes tree}},'' \emph{\BIBforeignlanguage{{English}}{{International
  Journal of Robotics Research}}}, vol.~{31}, no. {2, SI}, pp. {216--235},
  {FEB} {2012}, {9th International Workshop on Algorithmic Foundations of
  Robotics (WAFR), Natl Univ Singapore, Singapore, SINGAPORE, DEC 13-15, 2010}.

\bibitem{hindiyeh2013dynamics}
R.~Y. Hindiyeh, ``Dynamics and control of drifting in automobiles,'' Ph.D.
  dissertation, PhD Thesis, Stanford University, Stanford, California, USA,
  2013.

\end{thebibliography}

\end{document}